\documentclass[10pt,twocolumn,letterpaper]{article} 

\usepackage{avss}
\usepackage{times}
\usepackage{epsfig}
\usepackage{graphicx}
\usepackage{amsmath}
\usepackage{amssymb}

\usepackage{times}
\usepackage{epsfig}
\usepackage{xcolor}
\usepackage{float}
\usepackage{bm}
\usepackage{subcaption}

\newif\ifcomments
\commentsfalse

\ifcomments
    \usepackage{ulem}
    \def\mc#1{{\color{blue} [\textbf{MC:} #1]}}
    
    \def\jlk#1{{\color{red} [\textbf{JLK:} #1]}}
    
    \def\picomment#1{{\color{magenta} [\textbf{PI:} #1]}}

\else 
    \def\mc#1{}
    
    \def\jlk#1{}
    
    \def\picomment#1{}
    
    \def\zledit#1{}
    
\fi

\usepackage[pagebackref=true,breaklinks=true,letterpaper=true,colorlinks,bookmarks=false]{hyperref}

\avssfinalcopy 

\def\avssPaperID{14} 

\ifavssfinal\fi
\begin{document}


\title{Estimating Distances Between People using a Single Overhead Fisheye Camera with Application to Social-Distancing Oversight}
      

\author{Zhangchi Lu, Mertcan Cokbas, Prakash Ishwar, Janusz Konrad\thanks{This work was supported by ARPA-E (agreement DE-AR0000944) and by Boston University Undergraduate Research Opportunities Program.
}\\
Department of Electrical and Computer Engineering, Boston University\\
8 Saint Mary's Street, Boston, MA 02215\\
{\tt\small [zhchlu, mcokbas, pi, jkonrad]@bu.edu}
}

\maketitle

\begin{abstract}
  Unobtrusive monitoring of distances between people indoors is a useful tool in the fight against pandemics.  A natural resource to accomplish this are surveillance cameras. Unlike previous distance estimation methods, we use a single, overhead, fisheye camera with wide area coverage and propose two approaches. One method leverages a geometric model of the fisheye lens, whereas the other method uses a neural network to predict the 3D-world distance from people-locations in a fisheye image. To evaluate our algorithms, we collected a first-of-its-kind dataset using single fisheye camera, that comprises a wide range of distances between people (1--58 ft) and will be made publicly available. The algorithms achieve 1--2 ft distance error and over 95\% accuracy in detecting social-distance violations.
   \vglue -0.5cm
\end{abstract}

\section{Introduction}

The general problem of depth/distance estimation in 3D world has been studied in computer vision from its beginnings. However, the narrower problem of estimating the distance between people has gained attention only recently. In particular, the COVID pandemic has sparked interest in inconspicuous monitoring of social-distance violations (e.g., less than 6ft) \cite{IPM_distance, track_and_estimate, CNN_distance, YOLO_3_distance,social_deep, Pose_estimation, KORTE_dataset}. 
A natural, cost-effective resource that can be leveraged to accomplish this goal are the
surveillance cameras widely deployed in commercial, office and academic buildings.
%



Recent methods developed for the estimation of 3D distance have typically used 2 cameras (stereo) equipped with either rectilinear \cite{stero_camera, distort_compensate} or fisheye \cite{equirect_images, disparity_offset} lenses. Stereo-based methods, however, require careful camera calibration (both intrinsic and extrinsic parameters) and are very sensitive to misalignments between cameras (translation and rotation) after calibration. Although methods have been proposed using single rectilinear-lens camera \cite{track_and_estimate, CNN_distance,social_deep, KORTE_dataset, Pose_estimation}, that do not suffer from stereo calibration and misalignment shortcomings, usually one such camera can cover only a fragment of a large space. While multiple cameras can be deployed, this increases the cost and complexity of the system.



In this paper, we focus on estimating the distance between people indoors using a \textit{single} overhead {\it fisheye} camera with 360$^\circ\times$180$^\circ$ field of view. Such a camera can effectively cover a room up to 2,000ft$^2$ greatly reducing deployment costs compared to multiple rectilinear-lens cameras. However, fisheye cameras introduce geometric distortions so methods developed for rectilinear-lens cameras are not directly applicable; the geometric distortions must be accounted for when estimating distances in 3D space.


We propose two methods to estimate the distance between people using a single fisheye camera. The first method leverages a fisheye-camera model and its calibration methodology developed by Bone \textit{et al.}~ \cite{geo_paper} to inverse-project location of a person from fisheye image to 3D world. This inverse projection suffers from scale (depth) ambiguity that we address by using a human-height constraint. Knowing the 3D-world coordinates of two people we can easily compute the distance between them. Unlike the first method based on camera geometry, the second method uses the Multi-Layer Perceptron (MLP) and is data-driven.
In order to train the MLP, we collected training data using a large chess mat. For testing both methods, we collected another dataset with people placed in various locations of a 72$\times$28-foot room. The dataset includes over 300 pairs of people with over 70 different distances between them. 
Unlike other inter-people distance-estimation datasets, our dataset, due to the large field of view, includes a wide range of distances between people (from 1 ft to 58 ft). We call this dataset {\it Distance Estimation between People from Overhead Fisheye cameras} (DEPOF). 

The main contributions of this work are: 

\begin{enumerate}

\item \textbf{We propose two approaches
for distance estimation between people using a \textit{single overhead fisheye camera}.} To the best of our knowledge no such approach has been developed to date.

\item \textbf{We created a fisheye-camera dataset for the evaluation of inter-people distance-estimation methods.} This is the first dataset of its kind that 
is publicly available at \href{https://vip.bu.edu/depof}{vip.bu.edu/depof}

\end{enumerate}

\section{Related Work}

In the last two years, spurred by the COVID pandemic, a number of methods have been developed to estimate distances between people indoors. Most of these methods comprise two key steps: the detection of people in an image, and estimation of the 3D-world distance between people.

In order to detect people/objects, some methods \cite{YOLO_3_distance,geo_flat} rely on YOLO, other methods \cite{CNN_distance,track_and_estimate} use Faster R-CNN and still other methods \cite{IPM_distance} use GMM-based foreground detection. However, this is not the focus of this paper; we assume that bounding boxes around people are available.



To estimate the distance between detected people, a number of approaches have emerged that use a single camera with rectilinear lens. Some approaches rely on typical dimensions of various body parts (e.g., shoulder width) \cite{Pose_estimation, KORTE_dataset}, while others perform a careful camera calibration \cite{track_and_estimate, social_deep, CNN_distance} to infer inter-person distances. Also, stereo-based methods (two cameras) have been recently proposed to estimate the distance to a person/object \cite{stero_camera, distort_compensate}, but they require very precise camera calibration and are sensitive to post-calibration misalignments.

Very recently, a single overhead fisheye camera was proposed to \textit{detect} social distance violations in buses (which is coarser goal than distance estimation), but no quantitative results were published \cite{Tsik2022}.
Fisheye-stereo is often used in front-facing configuration for distance estimation in autonomous navigation \cite{equirect_images, disparity_offset}, but recently it was proposed in overhead configuration for person re-identification indoors based on location rather than appearance \cite{geo_paper}.
To accomplish this, the authors developed a novel calibration method to determine both intrinsic and extrinsic fisheye-camera parameters. We leverage this study to calibrate our {\it single} fisheye camera and we use a geometric model developed therein.

In terms of benchmark datasets for estimating distances between people, Epfl-Mpv-VSD,  Epfl-Wildtrack-VSD, OxTown-VSD \cite{Pose_estimation} and KORTE \cite{KORTE_dataset} are prime examples. Out of them only Epfl-Mpv-VSD and KORTE include some indoor scenes. More importantly, however, all of them have been collected with rectilinear-lens cameras, and are not useful for our study. In fact, the first three datasets have been derived by Aghaei \textit{et al.}~\cite{Pose_estimation} from Epfl-Mpv\cite{epfl_mpv}, Epfl-Wildtrack \cite{epfl_wildtrack} and OxTown \cite{Oxtown} datasets that are designed for people detection and tracking. In contrast, our datasets have been specifically designed for the estimation of distances between people in an indoor, large-space setting under a variety of occlusion scenarios.




\section{Methodology} \label{sec:methodology}

\begin{figure}[!t]
  \centering
  \vglue 0cm
  \includegraphics[width=0.45\textwidth]{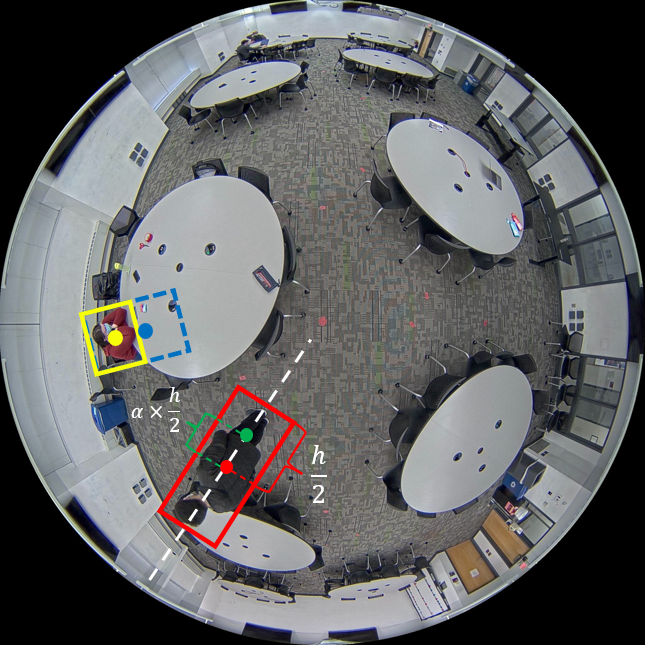}
  \caption{Field of view from an Axis M3057-PLVE camera mounted on the ceiling of a 72$\times$28 ft$^2$ classroom and illustration of height adjustment (see Section~\ref{ssec:heightadj} for details).}
  \label{fig:classroom_view}
  \vglue -0.5cm
\end{figure}

We focus on large indoor spaces monitored by a single, overhead, fisheye camera. An example of an image captured in this scenario is shown in Fig.~\ref{fig:classroom_view}. We propose two methods to measure the distance between two people visible in such an image. One method uses a geometric model of a previously calibrated camera while the other makes no assumptions about the camera and is data-driven. Although these methods are well-known, we apply them in a unique way to address the distance estimation problem using a {\it single} fisheye camera.

In this work, we are not concerned with the {\it detection} of people; this can be accomplished by any recent method developed for overhead fisheye cameras \cite{Tamura_et_al, Li_et_al, RAPiD}. Therefore, we assume that tight bounding boxes around people are given. Furthermore, we assume that the center of a bounding box defines the location of the detected person.

Let $\bm{x}_A, \bm{x}_B \in \mathbb{Z}^2$ be the pixel coordinates of bounding-box centers for person $A$ and person $B$, respectively. Given a pair $(\bm{x}_A,\bm{x}_B)$, the task is to estimate the 3D-world distance between people captured by the respective bounding boxes. Below, we describe two methods to accomplish this.

\subsection{Geometry-based method}


In this approach, to estimate the 3D-world distance between two people we adopt the {\it unified spherical model} (USM) proposed by Gayer and Danilidis \cite{Geyer01} for fisheye cameras and a calibration methodology to find this model's parameters developed by Bone \textit{et al.} \cite{geo_paper}. This model, derived in \cite{geo_paper}, enables computation of an inverse mapping from image coordinates to 3D space as described next.

\begin{figure}[!htb]
  \centering
  \includegraphics[width=0.45\textwidth]{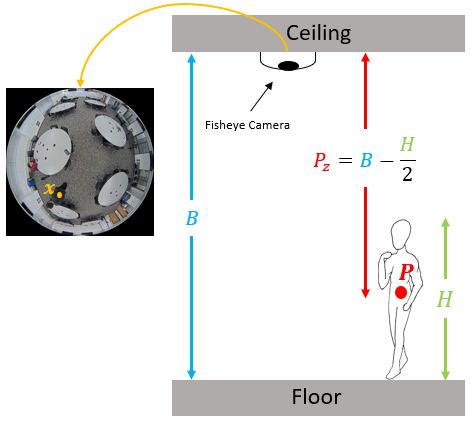}
  \caption{Illustration of the relationship between $P_z$ and person's height $H$. \label{fig:p_z_illustration}}
\end{figure}

Consider the scenario in Fig.~\ref{fig:p_z_illustration} where the center of the 3D-world coordinate system is at the optical center of a fisheye camera mounted overhead at height $B$ above the floor and a person of height $H$ stands on the floor. Let a 3D-world point $\bm{P}=[P_x,P_y,P_z]^T\in \mathbb{R}^3$ be located on this person's body at half-height and let $\bm{P}$ 
appear at 2D coordinates $\bm{x}$ in the fisheye image.

Bone {\it et al.} \cite{geo_paper} showed that the 3D-world coordinates $\bm{P}$ can be recovered from $\bm{x}$ with knowledge of $P_z$ and a 5-vector of USM parameters ${\bm{\omega}}$ via a non-linear function $G$:
\begin{equation} \label{eqn:G}
  \bm{P} = G(\bm{x},P_z; {\bm{\omega}}),
\end{equation}
%
%
In order to estimate ${\bm{\omega}}$, an automatic calibration method using a moving LED light was developed in \cite{geo_paper}.
In addition to $\bm{\omega}$, the value of $P_z$ is needed since this is a 2D-to-3D mapping. However, based on Fig.~\ref{fig:p_z_illustration} we see that $P_z=B-H/2$.


In practice, we can only get a pixel-quantized estimate $\widehat{\bm{x}}$ of $\bm{x}$ from which we can compute an estimate $\widehat{\bm{P}}$ of $\bm{P}$ using (\ref{eqn:G}). Let $\widehat{\bm{P}}_A$ and $\widehat{\bm{P}}_B$ denote the estimated 3D-world coordinates of person $A$ and person $B$, respectively, based on the centers of their bounding boxes $\widehat{\bm{x}}_A$ and $\widehat{\bm{x}}_B$. Then, we can estimate the 3D-world Euclidean distance $\widehat{d}_{AB}$ between them via:
%
%
\begin{equation} \label{eqn:geo_based_final_estimation}
  \widehat{d}_{AB} = ||\widehat{\bm{P}}_A-\widehat{\bm{P}}_B||_2.
\end{equation}

\subsection{Neural-network approach}

In this approach, we train a neural network to estimate the distance between person $A$ and person $B$. Since the distance between two points in a fisheye image is invariant to rotation, we pre-process locations $\bm{x}_A$ and $\bm{x}_B$ before feeding them into the network. First, we convert $\bm{x}_A$ and $\bm{x}_B$ to polar coordinates: $\bm{x}_A \rightarrow(r_A,\theta_A)$ and $\bm{x}_B\rightarrow (r_B,\theta_B)$, where $r_\cdot$ denotes radius and $\theta_\cdot$ denotes angle. Then, we compute the angle between normalized locations as follows:
\begin{equation}
  \theta := (\theta_A - \theta_B) \textrm{ mod } \pi.
\end{equation}
Note that by its definition, $0\le\theta\le\pi$. Finally, we form a feature vector associated with locations $\bm{x}_A$ and $\bm{x}_B$ as follows: $\bm{V} = [r_{A},r_{B},\theta]^T$. 
We chose a regression Multi-Layer Perceptron (MLP) to estimate the 3D-world distance between people (in lieu of a CNN) since the input vector is a 3-vector with no required ordering of coordinates for which convolution would be beneficial. 
%
%
We collected a training set of images, where for each vector $\bm{V}$ we know the ground-truth distance $d_{AB}$, and trained the MLP, $F: \mathbb{R}^3  \mapsto \mathbb{R} $, as a regression model that performs the following mapping:
\begin{equation} \label{eqn:F}
  \widehat{d}_{AB} = F(\bm{V}).
\end{equation}
In training, we used the mean squared-error (MSE) loss:
\begin{equation} \label{eqn:loss_function}
  \mathcal{L} = \frac{1}{M} \sum_{i=1}^{M} || \widehat{d}_{AB_{i}} - d_{AB_{i}} ||^2,
\end{equation}
where $M$ is the batch size.

\subsection{Person's height adjustment}
\label{ssec:heightadj}


While the geometry-based approach can be tuned for specific height of a person through $P_z$ (\ref{eqn:G}), the neural-network approach would require 
%
a training dataset having a large number of annotated examples at multiple heights.
%
%
Since this is extremely labor intensive, we train the MLP at a single height of 32.5 inches (details in Section~\ref{sec:training data info}) which corresponds to one-half of $H =$ 65 in, an average person's height. Recall that we assume a person's location in the image is the center-point of the person's bounding box.
%
%
Therefore, for a standing, fully-visible 65-inch person this center-point matches the 32.5-inch training height well. However, there would be a mismatch for people of other heights or when a person is partially occluded, for example by a table. In the latter case, the detected bounding box would be above the table and so would be the bounding-box center. To compensate for this height mismatch between the training and testing data, we 
propose a test-time adjustment in the MLP approach.

This height adjustment can be thought of as lowering the center-point of a person in terms of pixel coordinates. An illustration of this idea is shown in Fig.~\ref{fig:classroom_view}
%
%
where the red point represents the center of the red bounding box and $h$ its height. In the process of height adjustment during test time, we move the {\it actual} center (red point) of the bounding box along the box's axis pointing to the center of the image (white-dashed line) to produce an {\it adjusted} center (green point).
This displacement is defined as $\alpha \times \frac{h}{2}$ and we consider a range of values for $\alpha$ (see Fig.~\ref{fig:height_adaptation_plot}).
A value of $\alpha > 0$ corresponds to moving the bounding-box center towards the image center, i.e., we reduce the height of a detected person.
%
%
\begin{figure*}[!ht]
  \centering
  \includegraphics[width=0.9\linewidth]{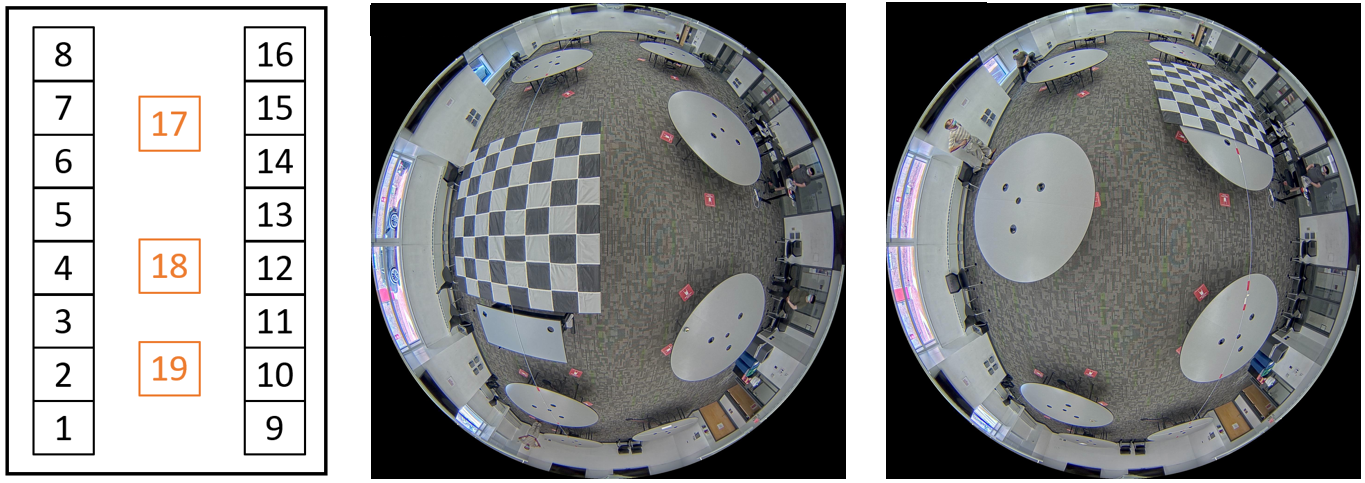}
  \leftline{\small\hskip 0.85cm (a) Layout of chessboard mats  \hskip 1.0cm (b) Chessboard mat at position \#4\hskip 1.5cm (c) Chessboard mat at position \#13}
  \caption{Illustration of chessboard-mat layout used for training the MLP model. 
  %
  %
  %
  %
  \label{fig:training_data}}
\end{figure*}

\section{Datasets} \label{sec:dataset info}

We introduce a unique dataset, {\it Distance Estimation between People from Overhead Fisheye cameras} (DEPOF) \footnote{\href{https:\\vip.bu.edu/depof}{vip.bu.edu/depof}} which was collected with Axis M3057-PLVE cameras at 2,048$\times$2,048-pixel resolution.

\subsection{Training dataset} \label{sec:training data info}


In order to train the MLP, we need ground-truth distance data. We placed a 9 ft $\times$ 9 ft chessboard mat on classroom tables of equal height (32.5 inches) in 19 different locations as shown in Fig.~\ref{fig:training_data}. The mat at location \#2 was carefully placed with sides parallel to those of the mat at location \#1 and forming a contiguous extension of mat \#1 (as if two mats were placed abutting each other). Similarly, the mat at location \#3 was placed contiguously with respect to the mat at location \#2 and so on until location \#8. The same procedure was repeated for locations \#9-\#16 with the mat at location \#9 carefully aligned with the mat at location \#1 and the distance between them carefully measured (121.5 inches). 
In order to provide ground-truth data in the center of camera's field of view, the mat was also placed directly under each of 3 cameras (locations \#17, \#18, \#19 shown in orange) with no alignment to mats at other locations.
\begin{figure}[!htb]
  \centering
  \includegraphics[width=0.35\textwidth]{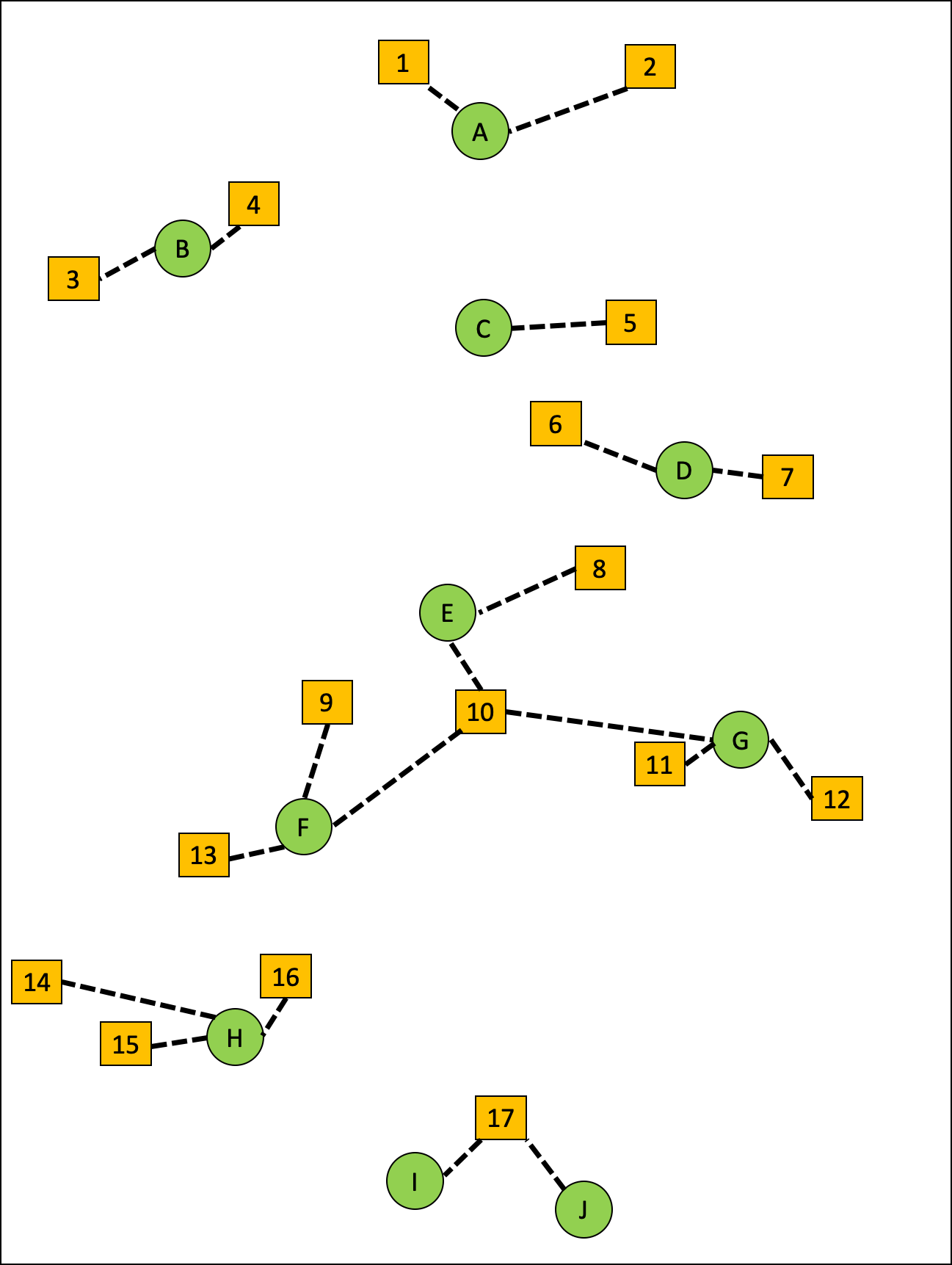}
  \caption{Spatial layout of locations in testing datasets. \label{fig:test_layout}}
  \vglue -0.4cm
\end{figure}

All the black/white corners of chessboard images were annotated, resulting in numerous $(\bm{x}_A,\bm{x}_B)$ pairs. Since we know that each square is of size 12.5 inches and the neighboring chessboards are abutting and aligned, 
we could accurately compute the 3D distances between physical-mat points corresponding to $\bm{x}_A$ and $\bm{x}_B$ 
The overall process can be thought of as creating a virtual grid with 12.5-inch spacing placed 32.5 inches above the floor of the classroom.



\subsection{Testing datasets}

In order to test both of the proposed approaches, we collected a dataset with people in a 72 ft $\times$ 28 ft classroom. First, we marked locations on the floor where individuals would stand  (Fig.~\ref{fig:test_layout}). We measured distances between all locations marked by a letter (green disk) 
which gives us ${10\choose 2} = 45$ distances which turned out to be distinct.
%
For locations marked by a number (yellow squares), we measured the distances along the dashed lines (20 distinct distances).
Using this spatial layout, we collected and annotated two sets of data. 

\begin{itemize}\itemsep 0em
\item{\bf Fixed-height dataset:} One person of height $H =$ 70.08 inches moves from one marked location to another and an image is captured at each location. This allows us to evaluate our algorithms on people of the same {\it known} height.


\item{\bf Varying-height dataset:} Several people of different heights stand at different locations in various permutations to capture multiple heights at each location. We use this dataset to evaluate sensitivity of our algorithms to a person's height variations.
\end{itemize}

In addition to the 65 distances ($45+20$), we performed $8$ additional measurements for the fixed-height dataset and $2$ additional measurements for the varying-height dataset.

Depending on
%
their location with respect to the camera, 
a person may be fully visible or partially occluded (e.g., by a table or chair). 
In order to understand the impact of occlusions on distance estimation, we grouped all the pairs in both testing datasets into 4 categories as follows:
%
%
\begin{itemize}\itemsep 0em
  \item Visible-Visible (V-V): Both people are fully visible.
  \item Visible-Occluded (V-O): One person is visible while the other one is partially occluded.
  \item Occluded-Occluded (O-O): Both people are partially occluded.
  \item All: All pairs, regardless of the occlusion status.
\end{itemize}

Table~\ref{tab:Dataset_stats} shows various statistics for both datasets: the number of pairs in each category, the number of distances measured and their range as well as the number of pairs with distance in three ranges: 0ft--6ft, 6ft--12ft and $>$12ft.



\begin{table}[htt]
\centering
\caption{Statistics of the testing datasets. \label{tab:Dataset_stats}}
\begin{tabular}{|c|c|c|}
\hline
 & Fixed- & Varying-\\
 & height & height\\
 & dataset & dataset\\
\hline
Number of V-V pairs & 35 & 100 \\
Number of V-O pairs & 32 & 126\\
Number of O-O pairs & 6 & 30\\
Number of All pairs & 73 & 256 \\
\hline
Number of distances & 73 & 67\\
\hline
Smallest distance (G to 11) & \multicolumn{2}{c|}{11.63 inches}\\
\hline
Largest distance (A to J) & \multicolumn{2}{c|}{701.96 inches}\\
\hline
Number of pairs: 0 ft to 6 ft & 25 & 45\\
Number of pairs: 6 ft to 12 ft & 15 & 73\\
Number of pairs: above 12 ft & 33 & 138\\
\hline
\end{tabular}
\end{table}

In both datasets, we found bounding boxes of people using a state-of-the-art people-detection algorithm for overhead fisheye images \cite{RAPiD} from which we computed their locations (bounding-box centers). To measure the real-world distances between people, we used a laser tape measure.

\section{Experimental Results} \label{sec:Results}

In this section, we evaluate both of the proposed methods. First, we describe the experimental setup. Then, we compare the methods in terms of distance estimation accuracy and evaluate the impact of person's height adjustment on performance. Finally, we assess the ability of both methods to detect social-distancing violations.



\subsection{Experimental setup}

In the  geometry-based approach, to learn parameters ${\bm{\omega}}$ of the inverse mapping $G$ (\ref{eqn:G}) we used the method described by Bone \textit{et al.}~\cite{geo_paper}. This method requires the use of 2 fisheye cameras, but is largely automatic and has to be applied only once for a given camera type (model and manufacturer). 
In the experiments, we used one camera at a time (3 cameras are installed in the test classroom -- locations \#17-\#19 in Fig.~\ref{fig:training_data}) and report the results only for the center camera due to space constraints. Results for other cameras are similar.

%



In the neural-network approach, we used an MLP with 4 hidden layers and 100 nodes per layer. In training, we used MSE loss (\ref{eqn:loss_function}) and Adam optimizer with 0.001 learning rate.

\subsection{Distance estimation evaluation} \label{sec:Distance Estimation Evaluation}

In Tables \ref{tab:Fixed Height Distance_Estimation_Result_table} and \ref{tab:Varying Height Distance_Estimation_Result_table}, we compare the performance of both methods when estimating the distance between people on the fixed-height and varying-height datasets, respectively. We report the mean absolute error (MAE) between the estimated and ground-truth distances:
\begin{equation} \label{eqn:MAE}
  \textrm{MAE} = \frac{1}{N} \sum_{i=1}^{N}|\widehat{d}_{AB_{i}} - d_{AB_{i}}|
\end{equation}
where $N$ is the number of pairs in the dataset while $\widehat{d}_{AB_i}$ and $d_{AB_i}$ are the estimated and ground-truth distances for the $i$-th pair $AB$, respectively.

It is clear from Table~\ref{tab:Fixed Height Distance_Estimation_Result_table} that the geometry-based approach using $H/2 =$ 35.04 inches (to compute $P_z$) consistently outperforms the same approach using $H/2 =$ 32.5 inches, which, in turn, significantly outperforms the neural-network approach trained on chess mats placed at the height of 32.5 inches. While it is not surprising that knowing a test-person's height of $H =$ 70.08 inches improves geometry-based method's accuracy, it is interesting that even assuming $H/2 =$ 32.5 inches the geometry-based approach significantly outperforms the MLP optimized for a fixed height of 32.5 inches during training.



\begin{table}[!htb]
\caption{Mean-absolute distance error between two people for the fixed-height dataset.}
\label{tab:Fixed Height Distance_Estimation_Result_table}
\centering
\begin{tabular}{|c|cccc|}
\hline
&
\multicolumn{4}{c|}{MAE [in]} \\
\hline
&  V-V & V-O & O-O & All \\
\hline
Geometry-based & 9.85 & 31.69 & 32.30 & 21.27 \\
($H/2 =$ 35.04 in) & & & &\\
\hline
Geometry-based & 12.20 & 39.90 & 42.64 & 26.84 \\
($H/2 =$ 32.5 in) & & & &\\
\hline
Neural network & 17.72 & 48.84 & 56.58 & 34.56\\
(trained on 32.5 in) & & & &\\
\hline
\end{tabular}
\end{table}



Similar performance trends can be observed in Table~\ref{tab:Varying Height Distance_Estimation_Result_table} for the varying-height dataset but with larger distance-error values than in Table~\ref{tab:Fixed Height Distance_Estimation_Result_table}. This is due to the fact that in the varying-height dataset people have different heights, so a selected parameter $H$ in the geometry-based algorithm or a training height in the neural-network algorithm cannot match all people's heights at the same time. 
%


\begin{table}[!htb]
\caption{Mean-absolute distance error between two people for the varying-height dataset.}
\label{tab:Varying Height Distance_Estimation_Result_table}
\centering
\begin{tabular}{|c|cccc|}
\hline
&
\multicolumn{4}{c|}{MAE [in]} \\
\hline
&  V-V & V-O & O-O & All \\
\hline
Geometry-based & 14.70 & 41.37 & 55.55 & 32.62\\
($H/2 = $ 35.04 in) & & & &\\
\hline
Geometry-based & 20.18 & 51.14 & 67.61 & 40.98\\
($H/2 = $ 32.5 in) & & & &\\
\hline
Neural network & 24.64 & 55.88 & 70.87 & 45.43\\
(trained on 32.5 in) & & & &\\
\hline
\end{tabular}
\end{table}


Note that for two fully-visible people of the same and {\it known} height (Table~\ref{tab:Fixed Height Distance_Estimation_Result_table}), the geometry-based algorithm has an average distance error of less than 10 inches. This error grows to about 21 inches for all pairs (visible and occluded). For people of different and {\it unknown} heights (Table~\ref{tab:Varying Height Distance_Estimation_Result_table}), the average error for pairs of fully-visible individuals (for $H/2 =$ 35.04 inches) is slightly above 1 ft and for all pairs it is less than 3 ft. While these might seem to be fairly large distance errors, one has to note that the distances between people are as large as 58.5 ft (702 inches).
%




In terms of the computational complexity, on an Intel(R) Xeon(R) CPU E5-2680 v4@2.40GHz both algorithms can easily support real time operation although the geometry-based algorithm is significantly faster. 
For example, suppose 3D-world distances are to be computed between all pairs of 100 image locations. The geometry-based algorithm can first map all pixel coordinates to 3D world coordinates (\ref{eqn:G}) and then compute the Euclidean distance for all ${100\choose 2}= $ 4,950 pairs. This, on average, takes 4 $\mu s$. The neural-network algorithm has to apply the MLP to all 4,950 pairs separately taking on average 949 $\mu s$.

\subsection{Impact of a person's height adjustment}
\label{sec: Impact of height adaptation trick}

As we discussed in Section~\ref{ssec:heightadj}, the centers of the detected bounding boxes may not reflect the true height of a person due to occlusions. In this context, we proposed a method to adjust a bounding-box center location during testing to compensate for the occlusion effect.

Here, we evaluate the impact of this height adjustment on each method's performance. For a fair comparison, we use $H/2 =$ 32.5 inches (table height) in the geometry-based method. Recall that the neural-network approach was trained on chess mats placed at this height.

\begin{figure}[!t] 
    \centering
    \begin{subfigure}[h]{0.41\textwidth}  
      \centering
      \includegraphics[width=0.95\textwidth]{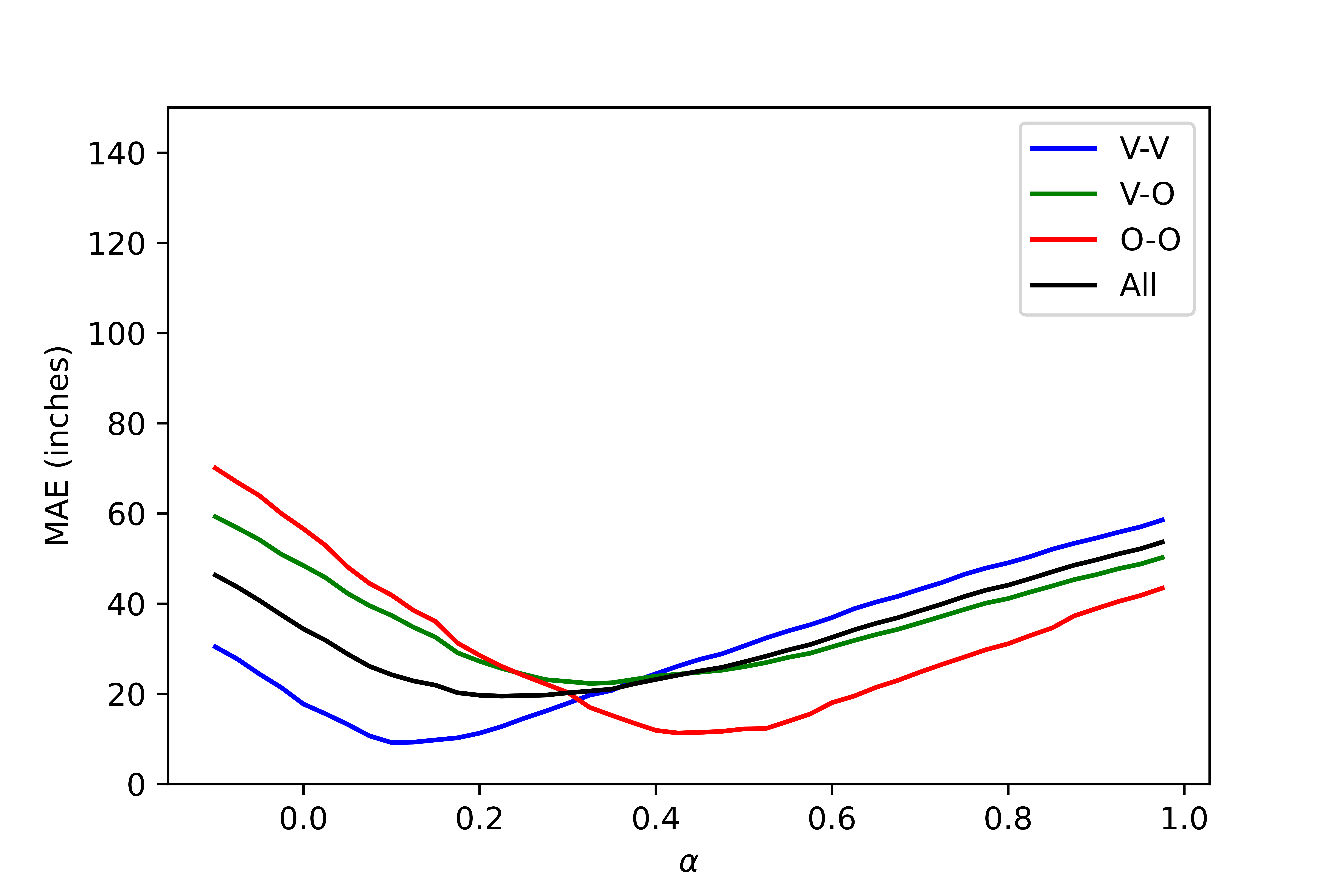}
      \caption{Neural-network algorithm, fixed-height dataset}
    \end{subfigure}
    \begin{subfigure}[h]{0.41\textwidth}  
      \centering
      \includegraphics[width=0.95\textwidth]{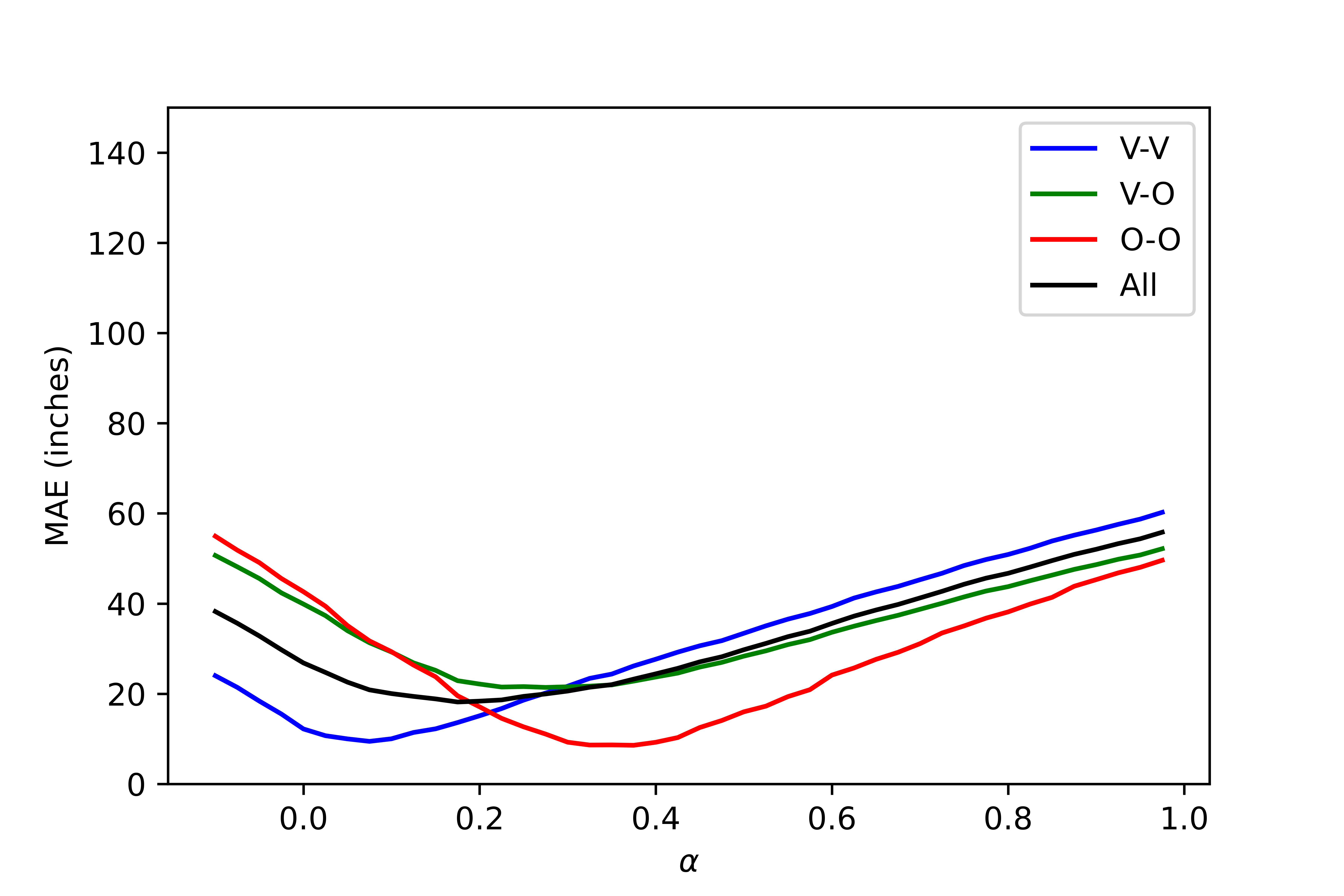}
      \caption{Geometry-based algorithm, fixed-height dataset}
    \end{subfigure}
    \begin{subfigure}[h]{0.41\textwidth}  
      \centering
      \includegraphics[width=0.95\textwidth]{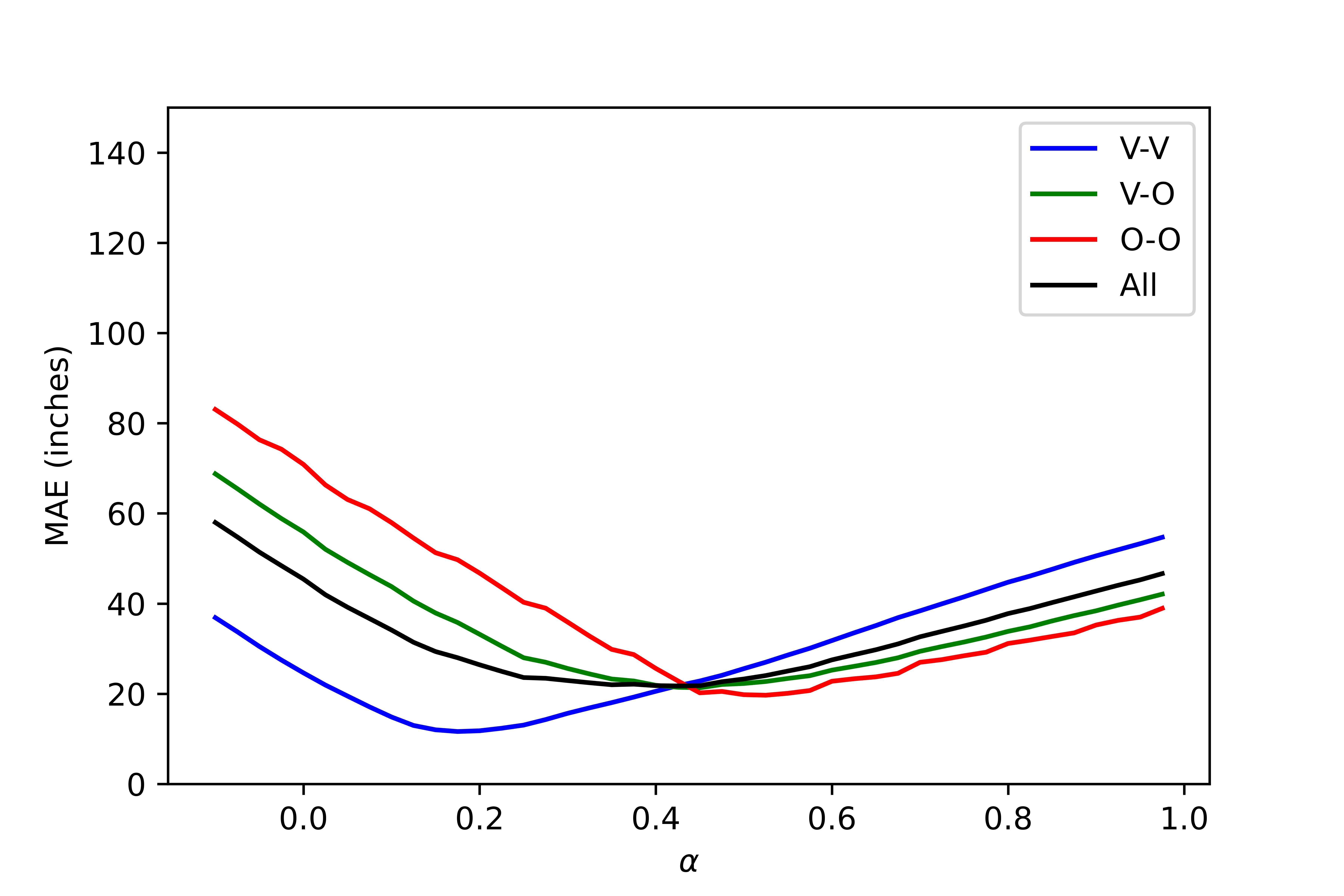}
      \caption{Neural-network algorithm, varying-height dataset}
    \end{subfigure}
    \begin{subfigure}[h]{0.41\textwidth}  
      \centering
      \includegraphics[width=0.95\textwidth]{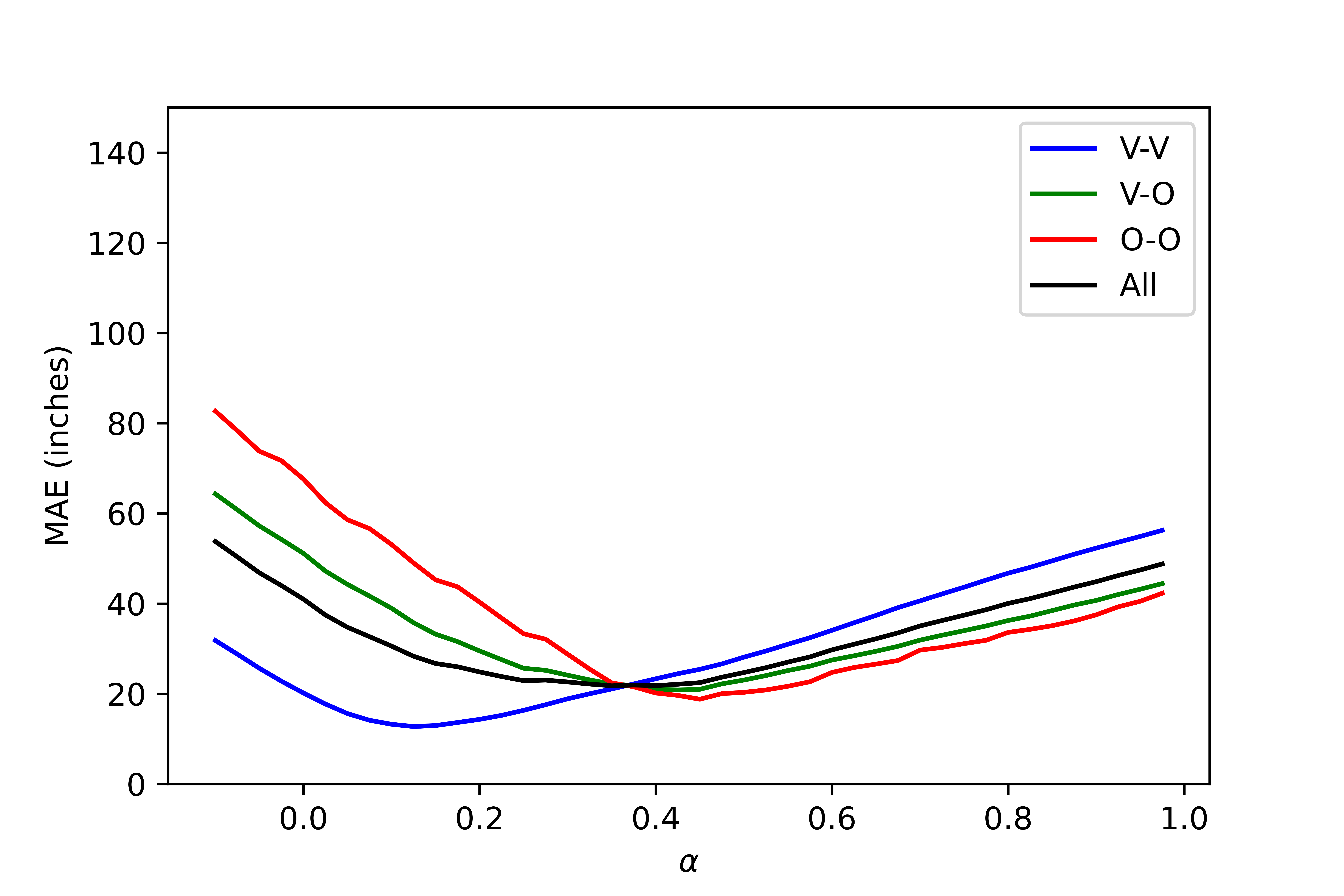}
      \caption{Geometry-based algorithm, varying-height dataset}
    \end{subfigure}
    \quad
    \caption{MAE for height adjustments: $-0.1\leq\alpha<1.0$.\label{fig:height_adaptation_plot}}
\end{figure}

The value of MAE as a function of height-adjustment parameter $\alpha$ is shown in Figs.~\ref{fig:height_adaptation_plot}(a) and \ref{fig:height_adaptation_plot}(b)
for the fixed-height dataset. When the true bounding-box centers are used ($\alpha=0$), the MAE for the neural-network approach and V-V pairs (blue line) is close to 20 inches. However, when the centers are lowered by about 10\% ($\alpha \approx$ 0.10), the MAE for V-V pairs drops to about 9 in. 
Similar trends can be observed in Figs.~\ref{fig:height_adaptation_plot}(c) and \ref{fig:height_adaptation_plot}(d)
for the varying-height dataset. If the true bounding-box centers are used, the MAE for V-V pairs is above 20 inches for the neural-network approach. However, when the centers are lowered by about 15\% ($\alpha \approx$ 0.15), the MAE for V-V pairs drops to around 12 inches. Analogous trends can be seen for other types of pairs and for all pairs, as well as for the geometry-based approach.

In Tables~\ref{tab:Fixed Height After Height Adaptation Distance_Estimation_Result_table} and \ref{tab:Varying Height After Height Adaptation Distance_Estimation_Result_table}, we show the lowest MAE values for each pair type along with the corresponding value of $\alpha$. The two methods perform quite similarly (except for O-O pairs in the fixed-height dataset on which the geometry-based method performs better). For example, for the fixed-height dataset in Table \ref{tab:Fixed Height After Height Adaptation Distance_Estimation_Result_table} MAE for the best $\alpha$ for V-V pairs drops to about 9 inches for both algorithms compared to 12-18 inches seen in Table~\ref{tab:Fixed Height Distance_Estimation_Result_table}. Overall, in both datasets, with the right choice of $\alpha$, the MAE is well below 2 ft, which can be argued to be a reasonable result considering that the inter-people distances in our dataset are up to 58.5 ft. 

Looking at Fig.~\ref{fig:height_adaptation_plot} and Tables~\ref{tab:Fixed Height After Height Adaptation Distance_Estimation_Result_table}-\ref{tab:Varying Height After Height Adaptation Distance_Estimation_Result_table}, one notes that MAE is minimized for much smaller values of $\alpha$ for V-V pairs ($\alpha =$ 0.08-0.17) than for O-O pairs ($\alpha =$ 0.32-0.51). 
%
This is because the majority of occlusions happen in the lower half of people's bodies in the testing dataset.
When a person is blocked in the bottom half, the bounding-box center radially shifts away from the image center. An example of this can be seen in Fig. \ref{fig:classroom_view}, where the person delineated by the yellow bounding box would have been delineated by the blue bounding box had there been no occlusion. Due to the occlusion, however, the bounding-box center shifts from the blue point to the yellow point. Therefore, the O-O pairs need to be compensated more than the V-V pairs, i.e., a higher value of $\alpha$ is needed.

\begin{table}[!htb]
\caption{Best mean-absolute distance error between two people for both methods using height adaptation for the fixed-height dataset. Under each MAE value, the corresponding optimum ``$\alpha$'' value used is written in parentheses. 
}
\label{tab:Fixed Height After Height Adaptation Distance_Estimation_Result_table}
\centering
\begin{tabular}{|c|cccc|}
\hline
&
\multicolumn{4}{c|}{MAE [in]}\\
&
\multicolumn{4}{c|}{($\alpha$)}\\
\hline
&  V-V & V-O & O-O & All\\
\hline
Geometry-based & 9.36 & 21.07 & 8.31 & 18.10 \\
($H/2 = $ 32.5in) & (0.08) & (0.28) & (0.32) & (0.18)\\
\hline
Neural network & 8.79 & 22.20 & 11.24 & 19.27 \\
(Trained on 32.5 in) & (0.12) & (0.33) & (0.42) & (0.26)\\
\hline
\end{tabular}
\end{table}

\begin{table}[!htb]
\caption{Best mean-absolute distance error between two people for both methods using height adaptation for the varying-height dataset. Under each MAE value, the corresponding optimum ``$\alpha$'' value used is written in parentheses.}
\label{tab:Varying Height After Height Adaptation Distance_Estimation_Result_table}
\centering
\begin{tabular}{|c|cccc|}
\hline
&
\multicolumn{4}{c|}{MAE [in]}\\
&
\multicolumn{4}{c|}{($\alpha$)}\\
\hline
&  V-V & V-O & O-O & All\\
\hline
Geometry-based & 12.76 & 20.49 & 18.66 & 21.48 \\
($H/2 = $ 32.5in) & (0.12) & (0.41) & (0.48) & (0.38)\\
\hline
Neural network & 11.62 & 21.30 & 18.60 & 21.47 \\
(Trained on 32.5 in) & (0.17) & (0.45) & (0.51) & (0.41)\\
\hline
\end{tabular}
\end{table}
In results reported thus far, the same value of $\alpha$ was used for both people in every pair. In the V-O and `All' categories, however, it could be advantageous to use different values of $\alpha$ for the visible and occluded person.
%
To verify this hypothesis, we 
applied $\alpha = 0.1$ to all visible bounding boxes and $\alpha = 0.5$ to all occluded bounding boxes in the fixed-height dataset. This $\alpha$ adjustment per person 
gave an MAE of 12.80 inches (down from 18.10 inches) for the geometry-based algorithm and 11.06 inches (down from 19.27 inches) for the neural-network approach. The corresponding MAE values for the varying-height dataset were: 13.97 inches (down from 21.48 inches) and 13.14 inches (down from 21.47 inches). Clearly, an automatic detection of body occlusions and a suitable adjustment of parameter $\alpha$ can further improve the distance estimation accuracy. 
This could be a fruitful direction for future work.


\subsection{Application in the context of social distancing}
\label{sec: Evaluation of Social Distancing Violation Detection}

One very practical application of the proposed methods is to detect situations when social-distancing recommendations (typically 6 ft) are being violated. This problem can be cast as binary classification: two people closer to each other than 6 ft are considered to be a ``positive'' case (violation takes place) whereas two people more than 6 ft apart are considered to be a ``negative'' case (no violation). To measure performance, we use Correct Classification Rate (CCR) and F1-score. Table~\ref{tab:Social Distance Result table} shows their values for both algorithms applied in this context to ``All'' pairs.  We report results for $\alpha = $ 0.5 which gives the lowest MAE for pairs with occlusions on the varying-height dataset (Table~\ref{tab:Varying Height After Height Adaptation Distance_Estimation_Result_table}).
On the fixed-height dataset, the neural-network approach slightly outperforms the geometry-based algorithm: by 1.37\% points in terms of CCR and by 0.63\% points in terms of F1-score. The methods perform identically on the varying-height dataset, achieving CCR value close to 95\% and F1-score close to 80\%. These results suggest that despite the presence of people of different heights both approaches achieve high enough CCR and F1-score values to be potentially useful in practice for the detection of social-distance violations in the wild. 

\begin{table}[!ht]
\caption{Social-distance violation detection results ($\alpha$=0.5).}
\label{tab:Social Distance Result table}
\vspace{-2ex}
\centering
\begin{tabular}{|c|cc|cc|}
\hline
&
\multicolumn{2}{c|}{Fixed-height} &
\multicolumn{2}{c|}{Varying-height}\\
&
\multicolumn{2}{c|}{dataset} &
\multicolumn{2}{c|}{dataset}\\
\hline
&  CCR & F1 & CCR & F1 \\
&  [\%] & [\%] & [\%] & [\%]\\
\hline

Geometry-based &  94.52 &  91.14 & 94.53 & 79.69\\
($H/2$=32.5in) & & & &\\
\hline
Neural network &  95.89 & 91.77 & 94.53 & 79.69\\
(Trained on 32.5in) & & & &\\
\hline
\end{tabular}
\vspace{-2ex}
\end{table}

\section{Concluding Remarks}

We developed two methods (the first of their kind) for estimating the distance between people in indoor scenarios based on a single image from a \textit{single} overhead fisheye camera. Demonstrating the ability to accurately measure the distance between people from a single overhead fisheye camera (with its wide FOV) has practical utility since it can decrease the number of cameras (and cost) needed to monitor a given area. A novel methodological contribution of our work is the use of a height-adjustment test-time pre-processing operation which makes the distance estimates resilient to height variation of individuals as well as body occlusions. We demonstrated that both methods can achieve errors on the order of 10-20in
for suitable choices of height-adjustment tuning parameters. We also showed that both of our methods can predict social distance violation with a high F1-score and accuracy. 

{\small
\bibliographystyle{ieee}
\bibliography{egbib}
}

\end{document}